\documentclass[lettersize,journal]{IEEEtran}
\usepackage{amsmath,amsfonts}
\usepackage{algorithmic}
\usepackage{array}
\usepackage[caption=false,font=normalsize,labelfont=sf,textfont=sf]{subfig}
\usepackage{textcomp}
\usepackage{stfloats}
\usepackage{url}
\usepackage{verbatim}
\usepackage{graphicx}
\usepackage{threeparttable} 
\usepackage{ragged2e}
\usepackage{algorithm}
\usepackage{listings}  
\usepackage{xcolor}   
\usepackage{listings}
\usepackage{multirow}
\usepackage{adjustbox}
\usepackage[numbers]{natbib}
\usepackage{hyperref}
\hypersetup{
    colorlinks=true,
    linkcolor=green,
    filecolor=magenta,      
    urlcolor=cyan,
    citecolor=cyan
} 
\definecolor{my_cyan}{RGB}{0,225,225}
\def\BibTeX{{\rm B\kern-.05em{\sc i\kern-.025em b}\kern-.08em
    T\kern-.1667em\lower.7ex\hbox{E}\kern-.125emX}}
\usepackage{balance}
\begin{document}
\title{LCD-Net: A Lightweight Remote Sensing Change Detection Network Combining Feature Fusion and Gating Mechanism}

\author{Wenyu Liu, Jindong Li, Haoji Wang, Run Tan, Yali Fu, Qichuan Tian$^{*}$
\thanks{Wenyu Liu, Haoji Wang, Run Tan, and Qichuan Tian are with the School of Electrical and Information Engineering, Beijing University of Civil Engineering and Architecture, Beijing, China. Email: \{2108110023028, 2108540023055, 2108110022010\}@stu.bucea.edu.cn, tianqichuan@bucea.edu.cn.}
\thanks{Jindong Li and Yali Fu are with the School of Artificial Intelligence, Jilin University, Changchun, China. Email: \{jdli21, fuyl23\}@mails.jlu.edu.cn.}
\thanks{*: Corresponding author.}
}
\markboth{Journal of \LaTeX\ Class Files,~Vol.~18, No.~9, September~2020}%
{How to Use the IEEEtran \LaTeX \ Templates}

\maketitle
\begin{abstract}
Remote sensing image change detection (RSCD) is crucial for monitoring dynamic surface changes, with applications ranging from environmental monitoring to disaster assessment. While traditional CNN-based methods have improved detection accuracy, they often suffer from high computational complexity and large parameter counts, limiting their use in resource-constrained environments.
To address these challenges, we propose a \underline{\textbf{L}}ightweight remote sensing \underline{\textbf{C}}hange \underline{\textbf{D}}etection \underline{\textbf{Net}}work (\textbf{LCD-Net} in short) that reduces model size and computational cost while maintaining high detection performance. LCD-Net employs MobileNetV2 as the encoder to efficiently extract features from bitemporal images. A Temporal Interaction and Fusion Module (TIF) enhances the interaction between bitemporal features, improving temporal context awareness. Additionally, the Feature Fusion Module (FFM) aggregates multi-scale features to better capture subtle changes while suppressing background noise. The Gated Mechanism Module (GMM) in the decoder further enhances feature learning by dynamically adjusting channel weights, emphasizing key change regions.
Experiments on LEVIR-CD+, SYSU, and S2Looking datasets show that LCD-Net achieves competitive performance with just 2.56M parameters and 4.45G FLOPs, making it well-suited for real-time applications in resource-limited settings. The code is available at \url{https://github.com/WenyuLiu6/LCD-Net}.
\end{abstract}


\section{Introduction}
\IEEEPARstart{R}{emote} sensing change detection methods involve analyzing remote sensing images acquired at different time points to identify and assess changes in surface features~\cite{2020_Remote_Sensing_Change_detection}. Remote sensing technology utilizes satellites, drones, and airborne platforms to collect information about the Earth's surface, enabling monitoring of environmental changes over large areas and with high temporal resolution. This technology not only provides rich spatial information but also reveals the dynamic evolution of surface features~\cite{2021_IEEE_Transactions_on_Geoscience_and_Remote_Sensing_Remote_sensing_image_change_detection_with_transformers,2024_International_Journal_of_Multidisciplinary_Sciences_and_Arts_Future_Horizons}, making it widely applicable in various fields such as environmental monitoring, urban planning, agricultural management, disaster assessment, and resource management~\cite{2022_Advanced_Materials_CNT,2023_Signal_End-to-end}.

In recent years, convolutional neural networks (CNNs) have become an important research focus in remote sensing image change detection (RSCD) due to their powerful feature extraction capabilities. Deep learning models can automatically learn multi-level feature representations, effectively capturing temporal changes between images. For instance, U-Net~\cite{2021_IEEE_Geoscience_and_Remote_Sensing_Letters_Unet} and its derivative models (such as FC-ef~\cite{2018_IEEE_FC}, FC-cat~\cite{2018_IEEE_FC}, FC-diff~\cite{2018_IEEE_FC}) show remarkable performance in multi-level feature fusion. These models efficiently combine features at different scales, improving detection accuracy and robustness. More complex architectures like DSIFN~\cite{2020_ISPRS_DSIFN} and SNUNet~\cite{2021_IEEE_Geoscience_and_Remote_Sensing_Letters_SNUNet-CD} further enhance detection performance through multi-layer feature fusion and attention mechanisms. These methods leverage high-level temporal features to detect changes between images, while low-level features capture detailed boundaries.
\begin{figure*}[ht]
    \centering
    \subfloat{\includegraphics[width=0.31\textwidth]{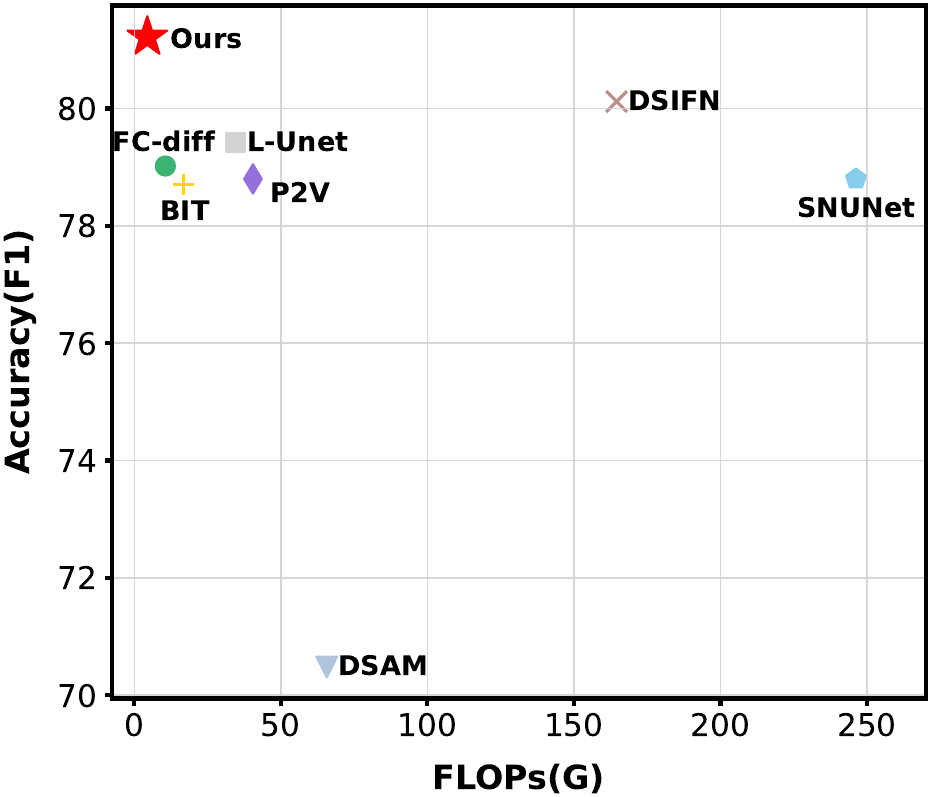}\label{fig:fig_01}}
    \hfill
    \subfloat{\includegraphics[width=0.31\textwidth]{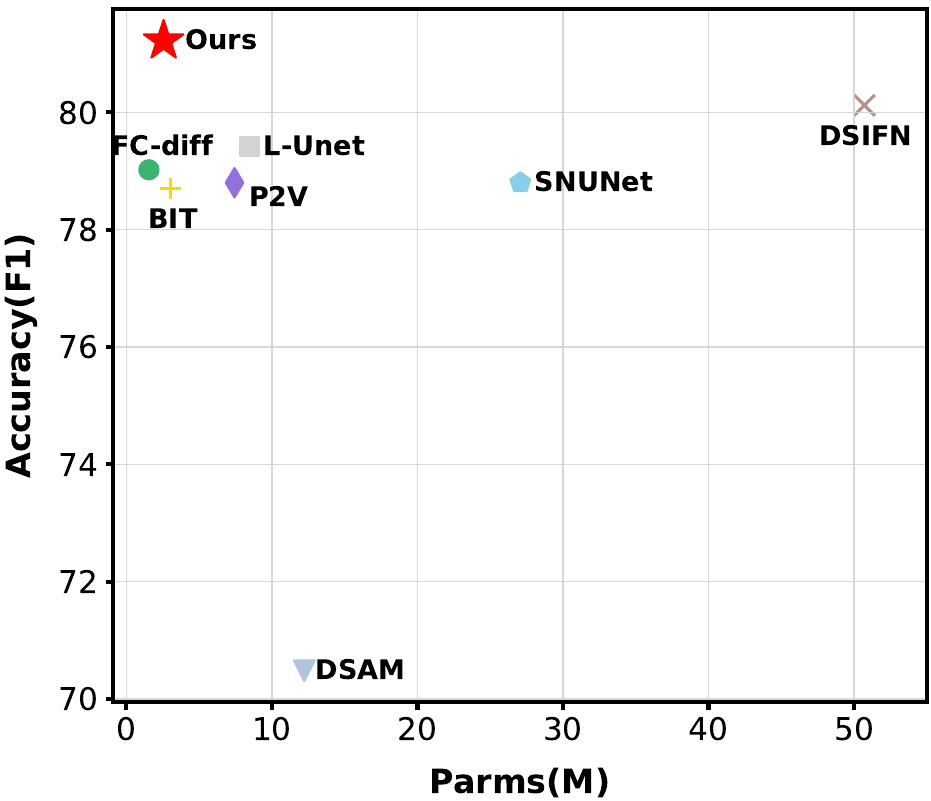}\label{fig:fig_02}}
    \hfill
    \subfloat{\includegraphics[width=0.31\textwidth]{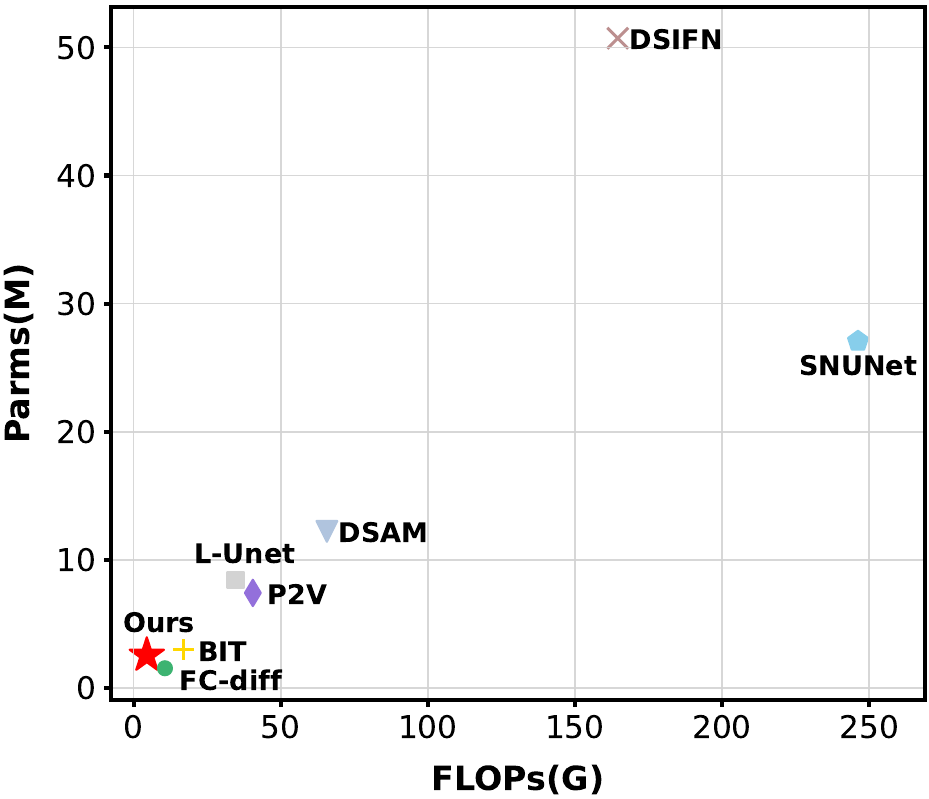}\label{fig:fig_03}}
    \caption{Comparison of model complexity on the SYSU dataset, illustrating FLOPs vs. F1-score, FLOPs vs. Params, and Params vs. F1-score for different methods.}
    \label{fig:fig_data_comparison}
\end{figure*}

Despite the notable advancements made by traditional CNN-based methods in remote sensing image change detection (RSCD), the practical implementation of these techniques is often hindered by their computational demands and limitations in effectively addressing the complexities of real-world scenarios. As the demand for real-time change detection increases, it becomes crucial to improve the efficiency of these models and enhance their robustness against challenging conditions. This leads us to identify two critical challenges in the current landscape of RSCD.

\textbf{Firstly}, as shown in Figure~\ref{fig:fig_data_comparison}, various RSCD methods proposed in recent years have performed well in terms of detection accuracy (F1-score). However, most models still exhibit high computational complexity and parameter counts, exceeding the capacity of resource-constrained environments. More complex models (such as SNUNet~\cite{2021_IEEE_Geoscience_and_Remote_Sensing_Letters_SNUNet-CD}) require substantial computational resources to achieve high detection accuracy, while some lightweight models (such as L-Unet~\cite{2021_IEEE_Transactions_on_Geoscience_and_Remote_Sensing_L-Net}) reduce computational costs and parameter counts but perform poorly in terms of detection accuracy. This high computational cost significantly limits the widespread application of these methods in practical situations, especially in environments with limited computational resources (such as common edge computing devices or mobile platforms). Thus, designing lightweight models with lower computational loads and fewer parameters while maintaining good detection accuracy has become a key issue that urgently needs to be addressed in the RSCD field. 

\begin{figure*}[ht]
    \centering
    \captionsetup[subfloat]{labelformat=empty} 
    \subfloat[\scriptsize image T1]{\includegraphics[width=0.18\textwidth]{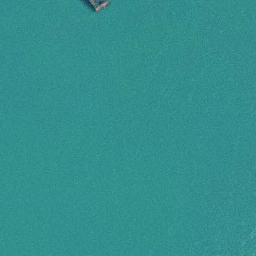}} 
    \hspace{0.005\textwidth}
    \subfloat[\scriptsize image T2]{\includegraphics[width=0.18\textwidth]{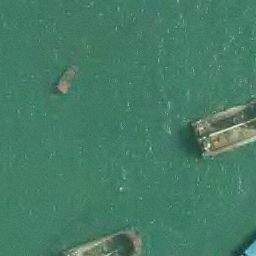}} 
    \hspace{0.005\textwidth}
    \subfloat[\scriptsize Real Change Map]{\includegraphics[width=0.18\textwidth]{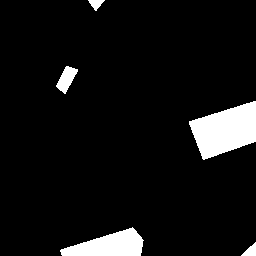}} 
    \hspace{0.005\textwidth}
    \subfloat[\scriptsize Without FFM]{\includegraphics[width=0.18\textwidth]{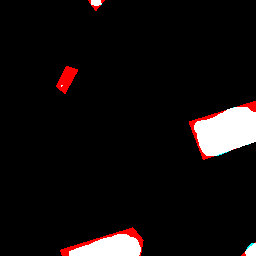}} 
    \hspace{0.005\textwidth}
    \subfloat[\scriptsize With FFM]{\includegraphics[width=0.18\textwidth]{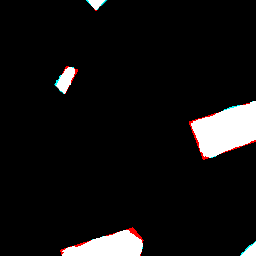}}
    \caption{This set of images is from the SYSU dataset. Among them, FFM represents the Feature Fusion Module. The displayed colors correspond to true positives (white), false positives (red), true negatives (black), and false negatives (blue).}
    \label{fig:fig_question2}
\end{figure*}

\textbf{Secondly}, interference from complex backgrounds and the coexistence of multi-scale changes pose critical challenges in the current change detection domain. Complex textures and similar structures in the background, as illustrated in Figure~\ref{fig:fig_question2}, such as water surface reflections and noise, obscure subtle target changes, making it difficult for detection methods to effectively distinguish between real changes and background interference. Meanwhile, traditional methods lack sensitivity to different scale changes, struggling to capture the differences between significant changes and subtle local variations. This weak response to multi-scale features significantly reduces detection accuracy, particularly in complex scenarios, leading to false positives and missed detections.

For the first challenge, we select the lightweight backbone network MobileNetV2~\cite{2023_Sustainability_MobileNetV2} as the encoder due to its excellent performance in feature extraction tasks and its minimal parameter count. However, the features extracted at a single moment often lack sufficient contextual information, making it difficult to comprehensively reflect the true nature of changes, which leads to inadequate feature extraction capability. To this end, we introduced a Temporal Interaction and Fusion Module (TIF) into the encoder. The module facilitates cross-temporal information interaction through the exchange of channel features, effectively integrating bi-temporal features and providing richer contextual information. This mechanism not only enhances the ability to perceive changes but also ensures that the model's parameter count remains unchanged.

For the second challenge, we propose the Feature Fusion Module (FFM) effectively addresses these challenges. By integrating multi-scale features across layers, the FFM enhances the model's sensitivity to multi-scale changes, especially in complex backgrounds, allowing it to better isolate background noise from real changes. Additionally, the FFM adopts a layer-by-layer approach, thoroughly extracting features from both significant and subtle changes. This enables the model to capture the fine differences between changes of various scales, improving its sensitivity and ability to distinguish changes. Furthermore, to further enhance the extraction capability of key features, we introduce a Gated Mechanism Module (GMM) in the decoding layer. This module dynamically adjusts learnable parameters to focus on feature learning in change areas, thereby improving the model's accuracy and robustness.

By incorporating the TIF, FFM, and GMM modules, our LCD-Net achieves competitive performance on the LEVIR-CD+ and SYSU S2Looking datasets, while maintaining a lightweight architecture with only 2.56M parameters and 4.45G FLOPs. 
Our main contributions could be summarized as follows:
\begin{itemize}
    \item{We propose a new lightweight remote sensing image change detection model, LCD-Net, which has a parameter size of only 2.56M and a computational complexity of 4.45G FLOPs. This model enhances the performance of change detection tasks while ensuring low computational consumption.}
    \item{We introduce a TIF that enhances feature expressiveness and semantic consistency through the sharing and interaction of cross-temporal features. To address the issue of complex background information obscuring subtle changes, we incorporate an FFM that deepens the integration of multi-scale features, thereby enhancing the model's sensitivity to subtle changes. Furthermore, we design the GMM to dynamically adjust channel weights, strengthening the contributions of important channels while reducing parameter redundancy.}
    \item{We conduct a series of experiments to validate the effectiveness and superiority of the proposed methods. Experimental results demonstrate that the LCD-Net model achieves better detection performance than nine state-of-the-art models across three benchmark datasets.}
\end{itemize}  

\section{Related work}
\subsection{Traditional Change Detection Methods}
Traditional change detection methods can be mainly divided into pixel-based and object-based approaches~\cite{2023_Engineering_Applications_of_Artificial_Intelligence_Dual-branch_network,7}. Pixel-based detection methods, such as image differencing~\cite{8}, image ratio~\cite{9}, correlation coefficient~\cite{10}, and change vector analysis (CVA)~\cite{11}, primarily focus on individual pixels by comparing the values of corresponding pixels in two or more images to determine whether changes have occurred in the region. These methods are easily affected by environmental noise and are suitable for change detection in simple scenes. In contrast, object-based methods utilize classification techniques~\cite{12} and clustering algorithms~\cite{13} to partition images into multiple objects and analyze the characteristics and spatial relationships of these objects to identify changed areas~\cite{14,15,16,17}. However, these traditional methods often struggle to detect subtle changes in complex backgrounds, leading to insufficient accuracy. Additionally, some spectral-based techniques like Principal Component Analysis (PCA)~\cite{2019_kurita_PCA} and Spectral Angle Mapper (SAM)~\cite{2017_shivakumar_SAM} have been used to exploit spectral differences for more effective change detection in multispectral images. These methods can improve performance in specific contexts, though they still face challenges in handling complex backgrounds.

\subsection{CNN-based Change Detection Methods}
In recent years, convolutional neural network (CNN)-based methods have emerged as a significant research direction in remote sensing image change detection (RSCD), primarily due to their powerful feature extraction capabilities \cite{18}. The U-Net architecture \cite{2021_IEEE_Geoscience_and_Remote_Sensing_Letters_Unet} has become popular for its effectiveness in multi-temporal image feature extraction and fusion. Various architectures have been developed to further enhance detection performance. For instance, the FC-ef, FC-cat, and FC-diff models \cite{2018_IEEE_FC}, proposed by researchers, integrate fully convolutional encoder-decoder structures \cite{2019_Sensors_Convolutional_encoder--decoder} with Siamese networks \cite{2017_IEEE_Geoscience_and_Remote_Sensing_Letters_Siamese}. This design allows for effective feature extraction by utilizing skip connections, capturing both high-level and low-level features for comprehensive change detection. The DSAM model \cite{2021_IEEE_Geoscience_and_Remote_Sensing_Letters_DSAM} incorporates the Convolutional Block Attention Module (CBAM) to differentiate between channel and spatial features at various stages, enhancing the model's focus on significant features. Additionally, the DSIFN model \cite{2020_ISPRS_DSIFN} employs attention mechanisms to fuse multi-level deep features with image difference features, significantly improving change map boundary integrity. Furthermore, the STANet model \cite{2020_Remote_Sensing_STANet} utilizes a spatiotemporal attention mechanism to extract more discriminative features, thus boosting detection accuracy. The P2V model \cite{2022_IEEE_Transactions_on_Image_Processing_P2V} addresses incomplete temporal modeling by employing two decoupled encoders, enhancing transformation identification in spatial and temporal dimensions. Lastly, the BIT model \cite{2021_IEEE_Geoscience_and_Remote_Sensing_Letters_BIT} leverages a bidirectional temporal convolution network to capture change patterns within time series data, adding robustness to change detection tasks.
\subsection{Feature Fusion and Gating Mechanisms}
In complex backgrounds, effectively capturing the differences between significant changes and subtle local variations remains a major challenge in change detection. To address these issues, the MapsNet model \cite{2022_International_Journal_mapsnet} effectively aggregates multi-level contextual information using a dual-stream fully convolutional network and multi-attention modules. The multi-directional fusion and perception network \cite{2021_Remote_sensing_xu} enhances the propagation of information within the network through a refined feature fusion module and adaptive weighted fusion strategies. The DGFNet model \cite{2021_Remote_Sensing_dgfnet} adopts an encoder-decoder architecture and reduces the semantic gap between different levels of features through a feature enhancement module and a dual gate fusion module, improving land cover classification performance. Furthermore, the MSGFNet model \cite{2024Remote_Sensing_msgfnet} combines the EfficientNetB4 model based on a Siamese network to effectively extract bi-temporal features and employs the MSGFM module to maintain boundary details and accurately detect changed targets. Building on the above studies, we propose feature fusion and gating mechanism modules to further enhance the accuracy and robustness of change detection.

\section{Methodology}
\subsection{Model Overview}
\begin{figure*}[t]
\centering
\includegraphics[width=1\textwidth]{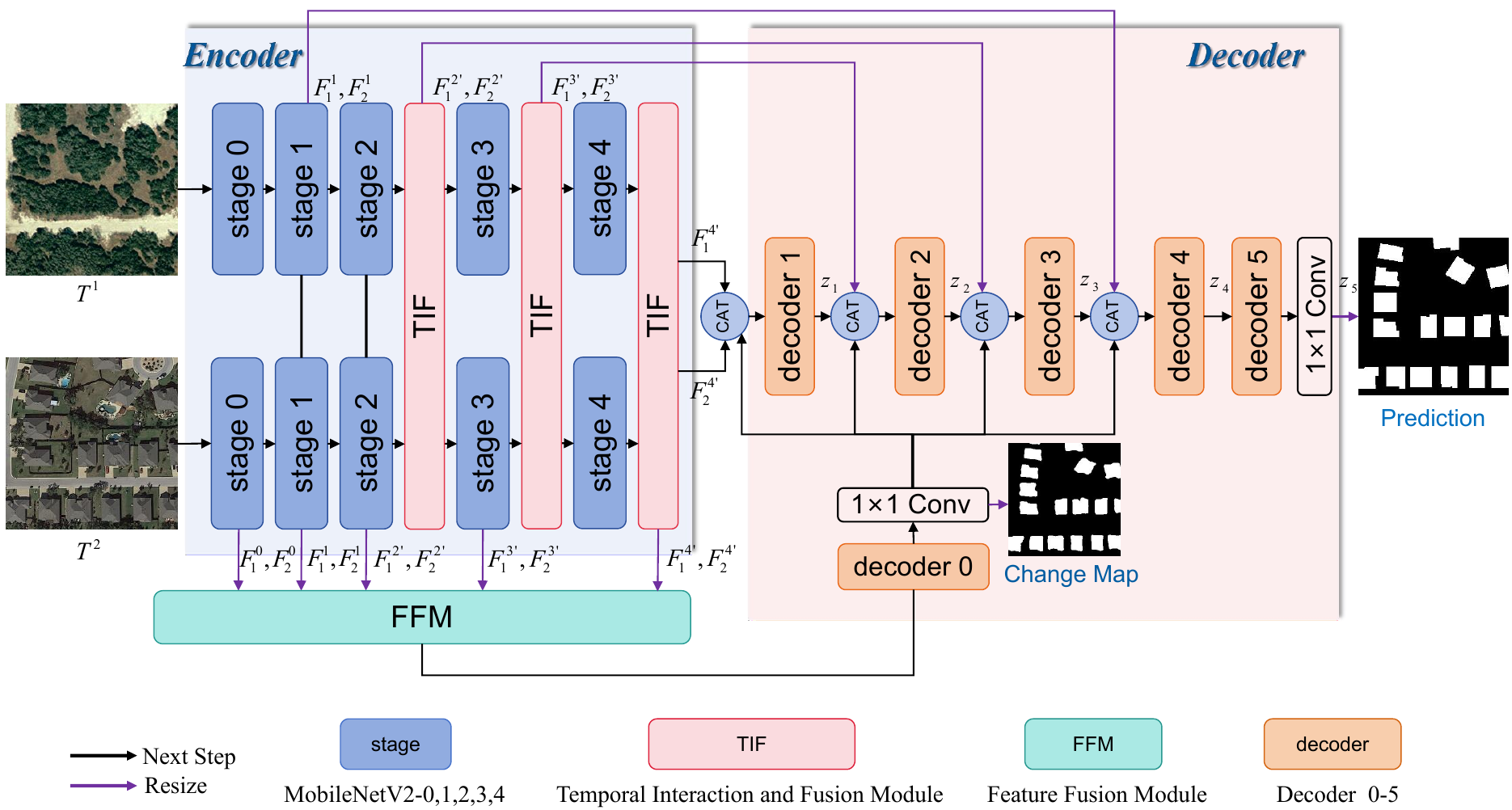}
    \caption{The proposed LCD-Net architecture.
The framework includes an encoder, Feature Fusion Module(FFM), and decoder. The encoder, based on MobileNetV2, extracts bi-temporal features enhanced by the TIF. The decoder upsamples features and uses the Gating Mechanism Module (GMM) to improve key feature extraction, producing the change detection map.}
    \label{fig:fig_framework}
\end{figure*}

In this section, we will provide a detailed overview of the LCD-Net model, as illustrated in Figure~\ref{fig:fig_framework}. The model consists mainly of three parts: the encoder, the feature fusion module (FFM), and the decoder. The encoder employs the MobileNetV2 network to extract dual-temporal features. To address the limitations of MobileNetV2~\cite{2023_Sustainability_MobileNetV2} in feature extraction capabilities, the model introduces a TIF. This module enhances the information flow between dual-temporal features through channel swapping, without increasing the model parameters, thereby improving feature extraction capability. After passing through the encoder, the dual-temporal remote sensing images T1 and T2 yield multi-level features $F_1^0, F_2^0, F_1^{4\textquotesingle}, F_2^{4\textquotesingle}$. These output features then enter the FFM, which achieves deep integration of multi-scale features while maintaining a small parameter size. Subsequently, the fused features are passed to the decoder. The decoder consists of multiple decoding layers that perform upsampling layer by layer and incorporate high-level features from the encoder for decoding. To further enhance the extraction capability of key features, a GMM is introduced in the decoding layers, dynamically adjusting the influence of each channel to improve the model's focus on critical information. Finally, the model generates the change detection map and prediction map through a 1×1 convolution operation. By integrating multi-scale feature extraction, channel exchange, and layer-wise decoding strategies, the model aims to improve the performance of change detection tasks while maintaining low computational consumption.

\subsection{Encoder}
\subsubsection{MobileNetV2 Network}
Traditional VGG networks~\cite{2018_VGG} and ResNet networks~\cite{2016_RESNet} perform well in feature extraction, but enhancing the feature extraction capability by increasing the depth of the convolution layer leads to significant computational overhead. In contrast, the MobileNetV2 network effectively reduces the computational complexity and number of parameters by introducing the deep separable linear convolution~\cite{32} and the Bottleneck structure~\cite{33}. In this paper, we use MobileNetV2 network as the coding layer and remove its global average pooling layer and the last fully connected layer. As shown in Figure 1, stage 0 to stage 4 in the encoder were used to extract the dual temporal features, and each stage contains a convolution layer with a stride of 2, so the feature map size decreases by half after each stage.
\subsubsection{Temporal Interaction and Fusion Module}
\begin{figure}
    \centering
    \includegraphics[width=0.3\textwidth]{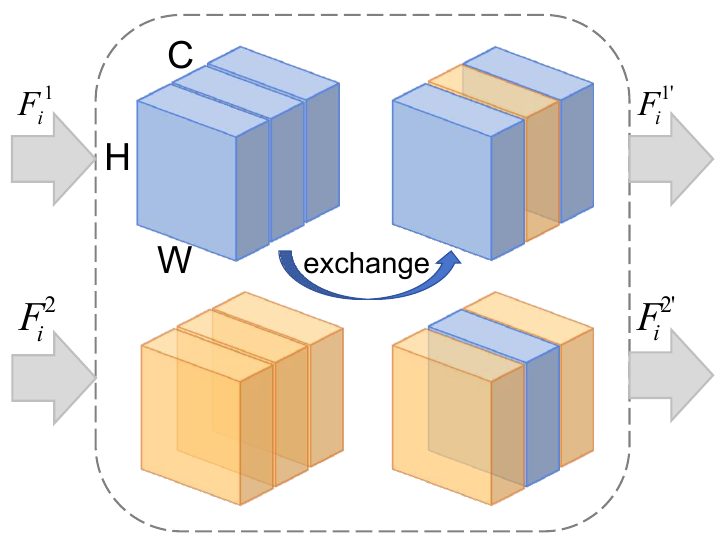}
        \caption{Illustration of the proposed Temporal Interaction and Fusion Module (TIF).}
        \label{fig:fig_TIF}
\end{figure}

To enhance the feature extraction capability of MobileNetV2, we introduce the TIF after stages 2 to 4. This module aims to improve the contextual awareness of dual-temporal images through the sharing and interaction of cross-temporal features. As shown in Figure~\ref{fig:fig_TIF}, the module exchanges features between the channels of dual-temporal images, allowing the model to capture temporal changes more effectively. By exchanging features between channels, the module enhances feature representation, improves semantic consistency, and mitigates the distribution differences between temporal features, thereby improving feature alignment. Moreover, this module does not require additional parameters or convolution operations, thus maintaining the lightweight nature of the model while improving its accuracy.

\begin{figure}
    \centering
    \includegraphics[width=\linewidth]{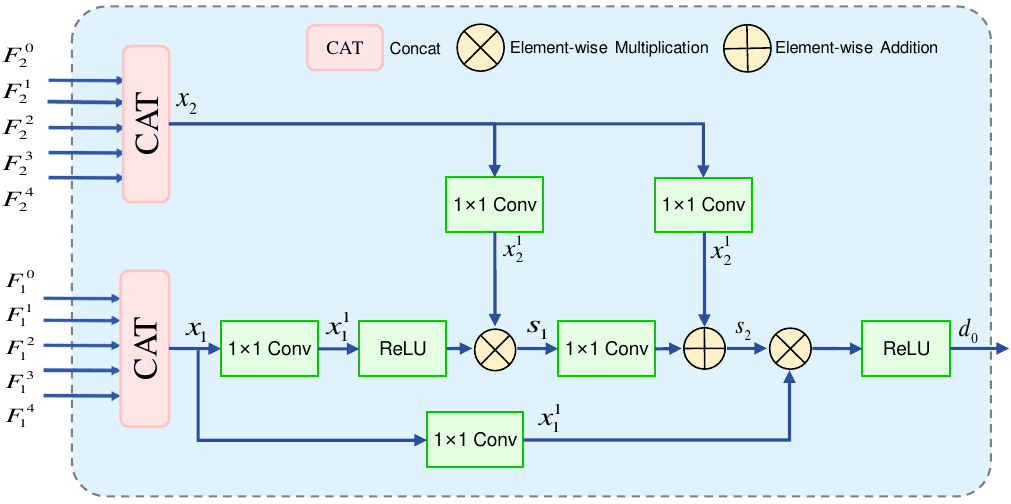}
    \caption{Illustration of the proposed Feature Fusion Module (FFM).}
    \label{fig:fig_FFM}
\end{figure}

\subsection{Feature Fusion Module}
In remote sensing image change detection (RSCD), subtle changes, especially in small vegetation areas and minor modifications to buildings, are difficult to capture. These change signals are often obscured by background information, affecting detection accuracy. Traditional methods that concatenate features directly tend to result in information redundancy, which makes it difficult to effectively capture these subtle changes. Inspired by the traditional cross-attention mechanism~\cite{34}, we propose the FFM, designed to optimize feature representation and information transmission, enhancing the model’s sensitivity to key features. This improves the detection of subtle changes while maintaining the lightweight nature of the model, as shown in Figure~\ref{fig:fig_FFM}. The FFM enhances feature extraction of subtle changes through an improved feature fusion strategy, boosting model performance in detecting changes. The specific process is as follows: The multi-level features $F_1^0, F_2^0 \cdots F_1^{4\textquotesingle}, F_2^{4\textquotesingle}$ obtained from the encoder are first concatenated (CAT) to form the preliminary feature combinations \( x_1 \) and \( x_2 \). Next, the concatenated features \( x_1 \) and \( x_2 \) undergo 1×1 convolution for channel adjustment, which reduces information redundancy and lowers computational complexity. After the convolution, the features \(x_1^1 \) are processed by the ReLU activation function and then multiplied with the feature \(x_2^1 \) to enhance the model's perception of subtle changes. This process amplifies subtle change features, enabling the model to more precisely differentiate between change regions and background information, This process can be represented by the following computation formula:
\begin{equation}
s_1=\operatorname{ReLU} \left(x_1^1\right) \times x_2^1
\end{equation}

The features $s_1$, after undergoing convolution, are combined with the original features $x_2^1$ through an addition operation. This ensures that the original feature information is retained while achieving a more comprehensive feature representation, thereby enhancing the model's ability to perceive change information. This process can be represented by the following computation formula:
\begin{equation}
s_2=\operatorname{Conv}\left(s_1\right)+x_2^1
\end{equation}

Finally, feature \( s_2 \) is multiplied element-wise with the original feature \(x_1^1 \) and optimized using the ReLU activation function to generate the output feature \( d_0 \). This process enhances the capture of change information and improves the precision of feature fusion. Through this design, the FFM achieves efficient feature integration and amplification of differences, improving the detection of subtle changes while reducing computational cost.

\begin{algorithm}
\caption{Feature Fusion Module (FFM)}
    \begin{lstlisting}
    # PyTorch-like Code
    import torch
    import torch.nn as nn
    
    class FFM(nn.Module):
        def __init__(self, in_planes1, in_planes2):
            super(FFM, self).__init__()
            self.conv1 = nn.Conv2d(in_planes1, in_planes2, kernel_size=1, stride=1, padding=0)
            self.relu = nn.ReLU(inplace=True)
    
        def forward(self, x1, x2):
            output_x1 = self.conv1(x1)
            output_x1 = self.relu(x1)
            output_x2 = self.conv1(x2)
            output_x = output_x1 * output_x2
            output_x = self.relu(output_x)
            output = output_x + x2
            output = output * x1
            output = self.relu(output)
            return output
    \end{lstlisting}
\label{alg:ffm}
\end{algorithm}

\subsection{Decoder}
\begin{figure}[b]
    \centering
    \includegraphics[width=\linewidth]{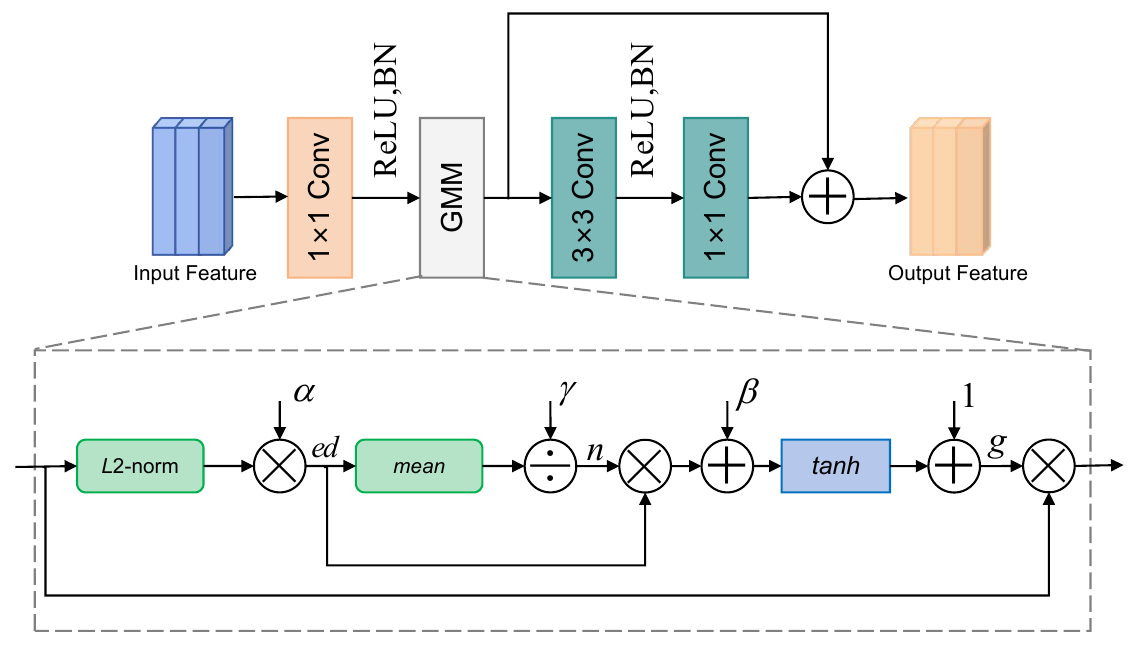}
    \caption{Illustration of the proposed encoder. The dashed box highlights the proposed Gated Mechanism Module (GMM).}
    \label{fig:fig_decoder}
\end{figure}

The decoder employs a hierarchical convolutional structure combined with the Gated Mechanism Module (GMM) to enhance the extraction of key features while maintaining a lightweight design, as shown in Figure~\ref{fig:fig_decoder}. First, a \(1 \times 1\) convolution is applied to reduce the channel number of the input features, eliminating information redundancy and decreasing computational complexity. The processed features are then dynamically adjusted using the GMM to optimize feature representation. Subsequently, a \(3 \times 3\) convolution is employed for spatial feature extraction and contextual information integration. To ensure stable gradient propagation in deep networks, a residual structure is introduced. Finally, \(1 \times 1\) convolution is used for feature decoding, producing the final feature map. The entire structure effectively extracts multi-scale features while maintaining low computational overhead.

\subsubsection{Gated Mechanism Module}
The GMM is the core component of the decoder, designed to enhance the expressive power of channel features while reducing the model parameters, as shown in Figure~\ref{fig:fig_decoder}. The GMM dynamically adjusts channel weights through three learnable parameters $(\alpha, \beta, \gamma)$. The parameter $\alpha$ is used to evaluate and aggregate the importance of each channel to effectively utilize channel information. The input features, processed through \textit{L2}-norm, are multiplied by parameter $\alpha$ to achieve effective channel aggregation. During the \textit{L2}-norm processing , a mechanism is introduced $\varepsilon\left(0 \leq \varepsilon \leq 10^{-5}\right)$ to prevent numerical instability. Compared to traditional fully connected layers, the normalization process of this module enables efficient modeling of channel relationships while reducing computational complexity. This process can be represented by the following computation formula:
\begin{equation}
ed=\alpha\left\|x_c\right\|_2=\alpha_c\left\{\left[\sum_{i=1}^H \sum_{j=1}^W\left(x_c^{i, j}\right)^2\right]+\varepsilon\right\}^{\frac{1}{2}}
\end{equation}
where x $\in R^{C \times H \times W}$ represents the input \(x\), \(x_c\) corresponds to the \(c\) channel of \(x\), \(x_c=\left[x_c^{i, j}\right]_{H \times W} \in R^{H \times W}\) , and \(\alpha_c\) denotes the learnable weight for the c channel.
The parameter \(\gamma\) is used to learn the competitive relationships between channels. The output feature 
\textit{ed} is divided by the parameter \(\gamma\) after mean normalization. This process effectively establishes competitive relationships between channels and reduces parameter redundancy. This process can be represented by the following calculation formula:
\begin{equation}
n=\frac{\gamma_c}{\sqrt{\text { mean }(e d)^2+\varepsilon}}
\end{equation}
where \(\gamma=\left[\gamma_1, \gamma_2, \cdots, \gamma_c\right]\) , $\gamma_c$ represents the learnable weights of channel \textit{c}. \(e_d\) is the feature calculated after mean normalization. The learnable parameter $\beta$ is used to amplify the contribution of important channels and reduce the impact of less important ones. By utilizing the gating function $1+\tanh (x)$, a nonlinear adjustment of the channel weights is achieved, which avoids the problem of gradient explosion, making the training process more stable. This process can be represented by the following mathematical formula:
\begin{equation}
g_c=1+\tanh \left(e d \cdot n+\beta_c\right)
\end{equation}
where \(\beta=\left[\beta_1, \beta_2, \cdots, \beta_c\right]\) , \(\beta_c\) the translation for the learnable bias of the \( c \) channel.

Finally, the output feature \textit{y} is obtained by performing an element-wise multiplication between the input tensor \textit{x} and the gating coefficient \textit{g}. Where\( \ g = \left[ g_1, g_2, \cdots ,g_c\right] \) , \(g_c \) the translation for the gating coefficient of the \( c \) channel. This process can be expressed by the following equation:
\begin{equation}
    y=x \cdot g
\end{equation}

\section{Experiments and Results}
\subsection{Experimental Setup}
The experiment is implemented on the PyTorch~\cite{35} platform, with an NVIDIA L20 GPU as the hardware environment. AdamW is chosen as the optimizer, with weight decay and learning rate set to 0.0025 and 0.0005, respectively. The model is trained for 100 epochs, and the loss function used was BCEWithLogitsLoss. All data are cropped to 256×256, with random Gaussian noise, salt-and-pepper noise, random cropping, and random rotation at certain angles applied. The experiments demonstrate that BCEWithLogitsLoss and AdamW accelerated the convergence speed of the network.
\begin{algorithm}[htbp]
\caption{Lightweigth Remote Sensing Change Detection Network (LCD-Net)}
\label{alg:LCD-Net_training}
\small
\begin{algorithmic}[1]
\REQUIRE Training dataset: $\mathcal{D}_{train} = \{T1, T2, label\}$
\ENSURE Anomaly scores for each graph $Score\_G$
\STATE \textbf{Initialize:} 
    \STATE (i) Data augmentation on images $T1$ and $T2$; 
    \STATE (ii) Load and fine-tune pre-trained MobileNetV2 as backbone for LCD-Net

\STATE \textbf{Training Phase}
\FOR{epoch = 1 \TO num\_epochs}
    \STATE epoch\_loss $\gets 0$
    \FOR{batch in train\_loader}
        \STATE Load $T1$, $T2$, $label$
        \STATE Zero gradients
        \STATE $preds \gets net(T1, T2)$
        \STATE $loss \gets \text{BCEWithLogitsLoss}(preds[0], label) + \text{BCEWithLogitsLoss}(preds[1], label)$
        \STATE Backpropagate and update model parameters
        \STATE epoch\_loss $\gets$ epoch\_loss + loss\_current\_epoch
        \STATE Update evaluator with binary predictions from sigmoid
    \ENDFOR
    \STATE Log metrics (IoU, Precision, Recall, F1-scores) and print epoch loss
\ENDFOR

\STATE \textbf{Validation Phase}
\FOR{batch in val\_loader}
    \STATE Load $T1$, $T2$, $label$
    \STATE $preds \gets net(T1, T2)$ \text{ with no gradient calculation}
    \STATE Update evaluator with binary predictions
\ENDFOR
\STATE Calculate final metrics for validation
\IF{new IoU $>$ best IoU} 
    \STATE Save best model
\ENDIF
\STATE Print best model metrics
\end{algorithmic}
\end{algorithm}

\subsection{Datasets and Evaluation Metrics}
\subsubsection{Datasets}
In this paper, experiments are conducted on three classic remote sensing image change detection datasets to evaluate the model's performance. The datasets used are LEVIR-CD+, SYSU, and S2Looking.

\textbf{LEVIR-CD+~\cite{36}:} The LEVIR-CD+ dataset is a large-scale dataset specifically designed for change detection tasks, consisting of 637 high-resolution image pairs, each with a size of 1024 x 1024 pixels. The dataset captures various types of building changes that occurred over a period of 5 to 14 years, primarily focusing on changes related to buildings, such as the growth and decline of structures. To annotate the change regions in these images, experts used binary labels, where “1” indicates a change and “0” indicates no change. In the entire dataset, there are 31,333 individual change samples, 21,442,501 change pixels, and 481,873,979 unchanged pixels. We use the standard data split, with 1,260 image pairs for training and 690 image pairs for testing. The original image pairs are cropped to a size of 256×256 pixels without overlap, ensuring a fair comparison with other state-of-the-art methods.

\textbf{SYSU~\cite{2021_IEEE_Geoscience_and_Remote_Sensing_Letters_DSAM}:} The SYSU dataset contains 20,000 ultra-high-resolution image pairs captured between 2007 and 2014 in Hong Kong, with each image having a size of 256×256 pixels. This dataset includes various types of changes, such as newly constructed urban buildings, suburban expansion, pre-construction groundwork, vegetation changes, road extensions, and marine developments. There are 286,092,024 changed pixels and 1,024,627,976 unchanged pixels. The SYSU dataset not only provides a large number of training samples but also presents a higher level of difficulty. Compared to the LEVIR-CD+ dataset, the imbalance ratio in the SYSU dataset is 1:3.58, and the larger dataset, along with subtle changes, increases the challenge for detection. We use the standard data split provided by the researchers on their website, which includes a total of 12,000 image pairs: 4,000 pairs for the validation set, 4,000 pairs for the training set, and 4,000 pairs for the test set.

\textbf{S2Looking~\cite{37}:} The S2Looking dataset consists of extensive side-view satellite images captured from different non-nadir angles, including approximately 65,920 annotated change instances and 5,000 pairs of dual-temporal images from rural areas. This dataset improves upon previous ones by providing a wider perspective, significant lighting variations, and the added complexity of rural photographs. It contains 66,552,990 changed pixels and 5,176,327,010 unchanged pixels, resulting in an imbalance ratio as high as 77.78. This makes it an extremely imbalanced dataset, where the ratio of foreground images (positive samples) to background images (negative samples) is 1:77.78. We use the standard data split provided by the researchers on their website, which includes 3,500 image pairs for the training set, 500 pairs for the validation set, and 1,000 pairs for the test set. The original image pairs are cropped to a size of 256×256 pixels without overlap to ensure a fair comparison with other state-of-the-art methods.
\subsubsection{Evaluation Metrics}
To evaluate the performance of the proposed method, several important metrics are employed, including Precision (PC), Recall (RC), F1-score (F1), Mean Intersection over Union (IoU), and Kappa coefficient (K). The calculation formulas are as follows:
\begin{equation}
    P C=\frac{T P}{T P+F P}
\end{equation}
\begin{equation}
    R C=\frac{T P}{T P+F N}
\end{equation}
\begin{equation}
    F 1=\frac{2 P C \cdot R C}{P C+R C}
\end{equation}
\begin{equation}
    O A=\frac{T P+T N}{T P+T N+F P+F N}
\end{equation}
\begin{equation}
    K=\frac{O A-P C}{T P+F P}
\end{equation}

\subsection{Comparison Experiments}

\subsubsection{Comparison of Model Performance}
To evaluate the performance of the proposed method, we compare it with state-of-the-art change detection methods under the same parameter settings during the training process. We selecte nine public change detection methods, including FC-cat~\cite{2018_IEEE_FC}, FC-ef~\cite{2018_IEEE_FC}, FC-diff~\cite{2018_IEEE_FC}, DSAM~\cite{2021_IEEE_Geoscience_and_Remote_Sensing_Letters_DSAM}, SNUNet~\cite{2021_IEEE_Geoscience_and_Remote_Sensing_Letters_SNUNet-CD}, L-Unet~\cite{2021_IEEE_Transactions_on_Geoscience_and_Remote_Sensing_L-Net}, BIT~\cite{2021_IEEE_Geoscience_and_Remote_Sensing_Letters_BIT}, DSIFN~\cite{2020_ISPRS_DSIFN}, and P2V~\cite{2022_IEEE_Transactions_on_Image_Processing_P2V}, as comparison subjects. We use accuracy, recall, F1-score, Kappa coefficient, and mean Intersection over Union as evaluation metrics for change detection accuracy. The experimental results are obtained by averaging the last five test results from multiple tests. Table~\ref{tab:tab1} presents the detection accuracy of the proposed method and the other nine methods on the LEVIR-CD+, SYSU, and S2Looking datasets. As shown in Table~\ref{tab:tab1}, the proposed method achieved the best performance on the LEVIR-CD+ dataset, with F1, IoU, and Kappa coefficients reaching 91.48\%, 84.30\%, and 91.03\%, respectively. Compared to existing methods, our method improved the F1-score by 0.41\%, the IoU by 0.69\%, and the Kappa coefficient by 0.43\%. On the SYSU dataset, the three metrics reach 81.22\%, 68.38\%, and 75.49\%, with improvements of 1.12\% in the F1-score, 1.54\% in IoU, and 1.34\% in Kappa. On the S2Looking dataset, the metrics reach 59.29\%, 62.00\%, and 56.8\%, with the F1-score improved by 1.46\%, IoU by 1.56\%, and Kappa by 1.23\%. These results indicate that the proposed method consistently outperformed other methods on the three datasets regarding the F1-score, Kappa coefficient, and mean IoU, which are the most important metrics.

\begin{table*}[h]
\caption{Comparison of the change detection metrics of different network models. (Color Convention:\textcolor{red}{best}, \textcolor{blue}{2rd best}) }
\centering
\renewcommand{\arraystretch}{1.3} 
\resizebox{0.99\textwidth}{!}{ 
\begin{tabular}{c|c c c c c c c c c c c}
\hline
Methods         &         & FC-cat & FC-ef & FC-diff & DSAM  & SNUNet & L-Unet & BIT   & DSIFN & P2V   & \textbf{{LCD-Net}} \\ 
\hline 
           & K(\%)   & 90.32  & 89.46 & 90.28   & 79.05 & 90.66  & 90.29  & 90.60 & 90.24 & 89.96 & \textcolor{red}{91.03}  \\
           & IoU(\%) & 83.16  & 81.79 & 83.10   & 66.65 & 83.70  & 83.11  & 83.61 & 83.03 & 82.59 & \textcolor{red}{84.30}   \\
LEVIR-CD+  & F1(\%)  & 90.81  & 89.98 & 90.77   & 79.98 & 91.13  & 90.78  & 91.07 & 90.73 & 90.47 & \textcolor{red}{91.48}   \\
           & RC(\%)  & 89.56  & 88.32 & 89.01   & 71.26 & \textcolor{red}{90.00}  & 89.80  & \textcolor{red}{90.88} & 88.64 & 89.07 & \textcolor{blue}{89.82}  \\
           & PC(\%)  & 92.09  & 91.72 & 92.59   & 91.14 & 92.28  & 92.58  & 92.09 & 92.91 & 91.91 & \textcolor{red}{93.20}   \\
\hline            
           & K(\%)   & 73.02  & 72.14 & 72.99   & 61.30 & 72.47  & 73.47  & 72.64 & 74.15 & 72.75 & \textcolor{red}{75.49}   \\
           & IoU(\%) & 65.40  & 64.55 & 65.32   & 54.42 & 65.01  & 65.87  & 64.87 & 66.84 & 65.02 & \textcolor{red}{68.38}   \\
SYSU       & F1(\%)  & 79.08  & 78.46 & 79.02   & 70.48 & 78.80  & 79.42  & 78.70 & 80.12 & 78.80 & \textcolor{red}{81.22}   \\
           & RC(\%)  & 75.49  & 75.53 & 75.01   & 70.87 & 76.85  & 75.77  & 74.11 & 78.51 & 74.46 & \textcolor{red}{80.56}   \\
           & PC(\%)  & 83.02  & 81.62 & 83.49   & 70.09 & 80.84  & \textcolor{red}{83.45}  & 83.88 & 81.40 & \textcolor{red}{83.68} & \textcolor{blue}{83.20}   \\
\hline 
           & K(\%)   & 54.70  & 46.03 & 52.22   & 48.48 & 57.04  & 55.87  & 55.94 & 57.34 & 51.65 & \textcolor{red}{58.93}   \\
           & IoU(\%) & 38.04  & 30.26 & 35.72   & 32.34 & 40.27  & 39.13  & 39.20 & 40.58 & 35.19 & \textcolor{red}{42.14}   \\
S2-Looking & F1(\%)  & 55.11  & 46.46 & 52.64   & 48.88 & 57.42  & 56.25  & 56.32 & 57.73 & 52.06 & \textcolor{red}{59.29}   \\
           & RC(\%)  & 54.89  & 40.95 & 50.03   & 41.98 & 54.50  & 53.20  & 53.75 & 57.39 & 56.48 & \textcolor{red}{56.81}  \\
           & PC(\%)  & 55.34  & 53.69 & 55.54   & 58.48 & 60.66  & 59.67  & 59.19 & 58.07 & 48.27 & \textcolor{red}{62.00}   \\
\hline
\end{tabular}
}
\label{tab:tab1}
\end{table*}

\begin{figure*}[ht]
    \centering
    \resizebox{\textwidth}{!}{ 
        \begin{tabular}{@{}ccccccccccccc@{}}
            \includegraphics[width=0.09\textwidth]{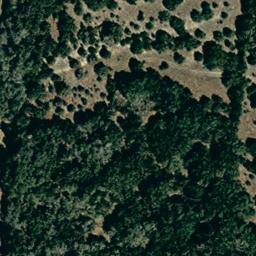} &
            \hspace{-4mm}
            \includegraphics[width=0.09\textwidth]{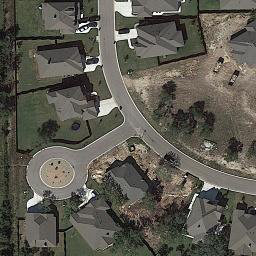} &\hspace{-4mm}
            \includegraphics[width=0.09\textwidth]{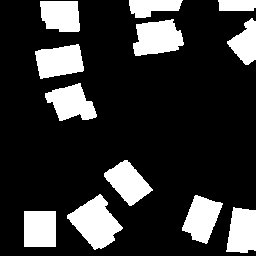} &\hspace{-4mm}
            \includegraphics[width=0.09\textwidth]{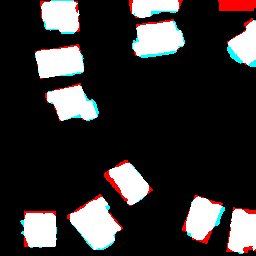} &\hspace{-4mm}
            \includegraphics[width=0.09\textwidth]{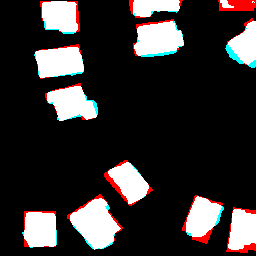} &\hspace{-4mm}
            \includegraphics[width=0.09\textwidth]{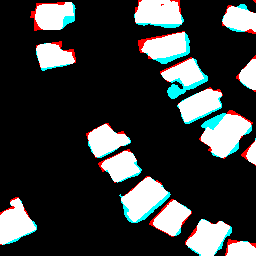} &\hspace{-4mm}
            \includegraphics[width=0.09\textwidth]{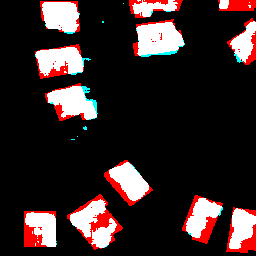} &\hspace{-4mm}
            \includegraphics[width=0.09\textwidth]{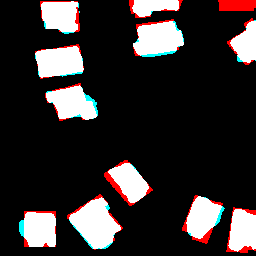} &\hspace{-4mm}
            \includegraphics[width=0.09\textwidth]{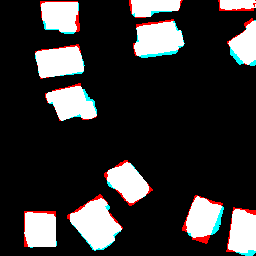} &\hspace{-4mm}
            \includegraphics[width=0.09\textwidth]{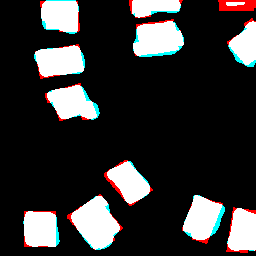} &\hspace{-4mm}
            \includegraphics[width=0.09\textwidth]{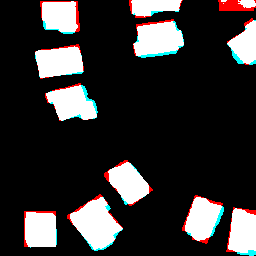} &\hspace{-4mm}
            \includegraphics[width=0.09\textwidth]{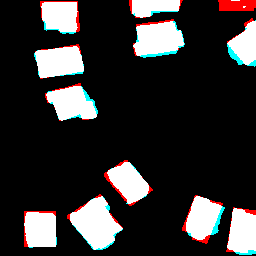} &\hspace{-4mm}
            \includegraphics[width=0.09\textwidth]{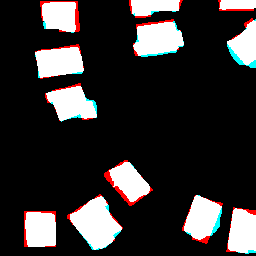} \\
            
            \includegraphics[width=0.09\textwidth]{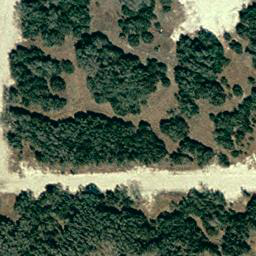} &\hspace{-4mm}
            \includegraphics[width=0.09\textwidth]{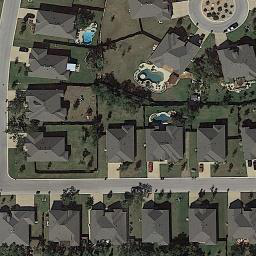} &\hspace{-4mm}
            \includegraphics[width=0.09\textwidth]{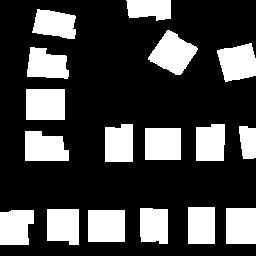} &\hspace{-4mm}
            \includegraphics[width=0.09\textwidth]{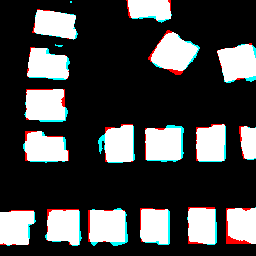} &\hspace{-4mm}
            \includegraphics[width=0.09\textwidth]{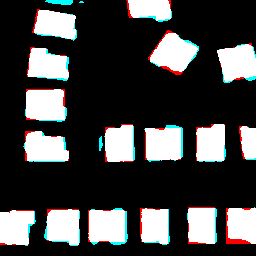} &\hspace{-4mm}
            \includegraphics[width=0.09\textwidth]{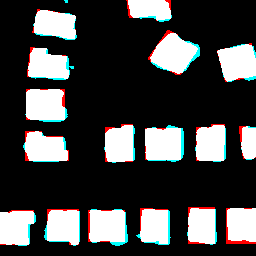} &\hspace{-4mm}
            \includegraphics[width=0.09\textwidth]{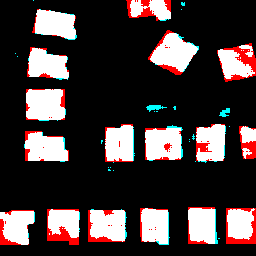} &\hspace{-4mm}
            \includegraphics[width=0.09\textwidth]{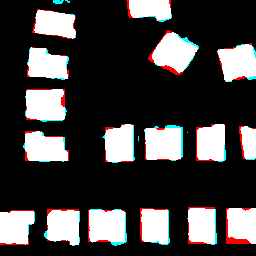} &\hspace{-4mm}
            \includegraphics[width=0.09\textwidth]{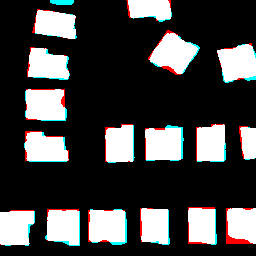} &\hspace{-4mm}
            \includegraphics[width=0.09\textwidth]{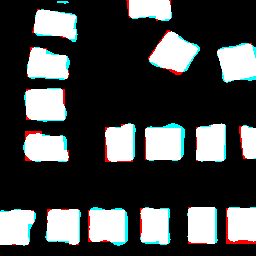} &\hspace{-4mm}
            \includegraphics[width=0.09\textwidth]{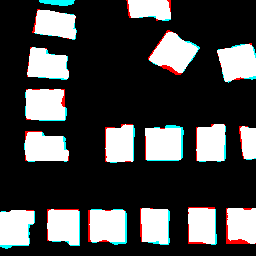} &\hspace{-4mm}
            \includegraphics[width=0.09\textwidth]{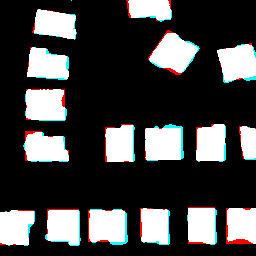} &\hspace{-4mm}
            \includegraphics[width=0.09\textwidth]{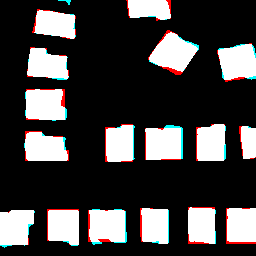} \\

            \includegraphics[width=0.09\textwidth]{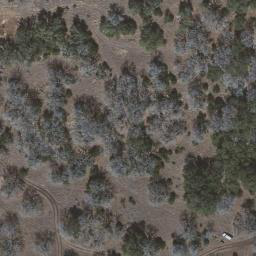} &\hspace{-4mm}
            \includegraphics[width=0.09\textwidth]{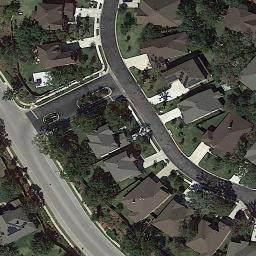} &\hspace{-4mm}
            \includegraphics[width=0.09\textwidth]{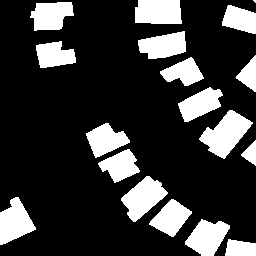} &\hspace{-4mm}
            \includegraphics[width=0.09\textwidth]{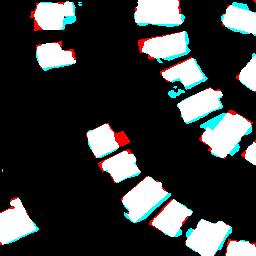} &\hspace{-4mm}
            \includegraphics[width=0.09\textwidth]{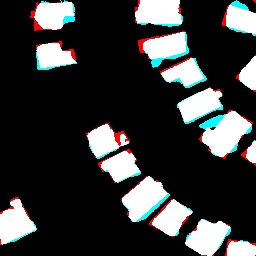} &\hspace{-4mm}
            \includegraphics[width=0.09\textwidth]{figs/LEVIR_12_15_FC-diff.png} &\hspace{-4mm}
            \includegraphics[width=0.09\textwidth]{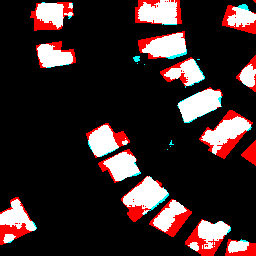} &\hspace{-4mm}
            \includegraphics[width=0.09\textwidth]{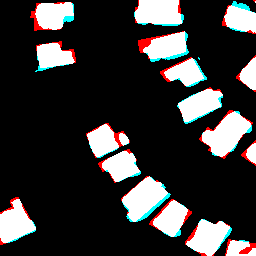} &\hspace{-4mm}
            \includegraphics[width=0.09\textwidth]{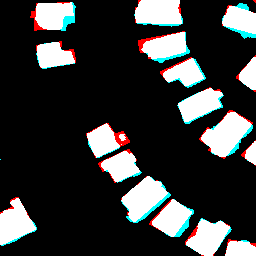} &\hspace{-4mm}
            \includegraphics[width=0.09\textwidth]{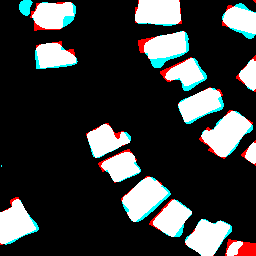} &\hspace{-4mm}
            \includegraphics[width=0.09\textwidth]{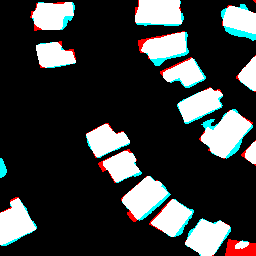} &\hspace{-4mm}
            \includegraphics[width=0.09\textwidth]{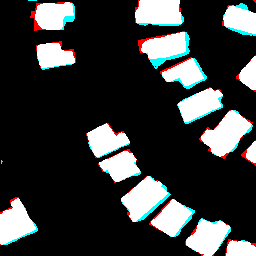} &\hspace{-4mm}
            \includegraphics[width=0.09\textwidth]{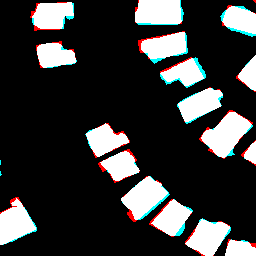} \\
            \small
            (a) & (b) & (c) & (d) & (e) & (f) & (g) & (h) & (i) & (j) & (k) & (l) & (m) \\
            \normalsize
        \end{tabular}
    }
    \caption{Qualitative analysis of experimental results on the LEVIR-CD+ dataset~\cite{36}. (a) T1 images, (b) T2 images, (c) Ground truth, (d) FC-cat, (e) FC-ef, (f) FC-diff, (g) DSAM, (h) SNUNet, (i) L-Unet, (j) BIT, (k) DSIFN, (l) P2V, (m) Ours. The displayed colors correspond to true positives (white), false positives (\colorbox{red!60}{red}), true negatives (black), and false negatives (\colorbox{my_cyan!40}{cyan}).}
    \label{fig:fig_LEVIR}    
\end{figure*}

\begin{figure*}[ht]
    \centering
    \resizebox{\textwidth}{!}{ 
        \begin{tabular}{@{}ccccccccccccc@{}}
            \includegraphics[width=0.09\textwidth]{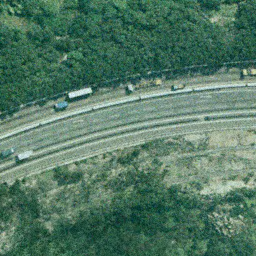} &\hspace{-4mm}
            \includegraphics[width=0.09\textwidth]{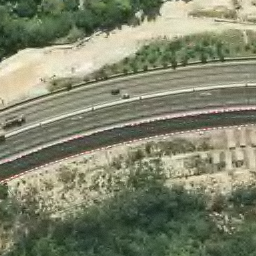} &\hspace{-4mm}
            \includegraphics[width=0.09\textwidth]{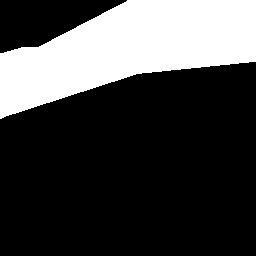} &\hspace{-4mm}
            \includegraphics[width=0.09\textwidth]{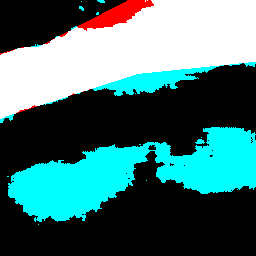} &\hspace{-4mm}
            \includegraphics[width=0.09\textwidth]{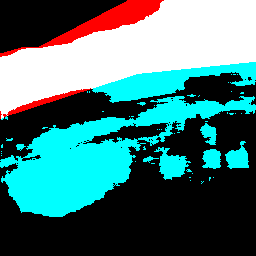} &\hspace{-4mm}
            \includegraphics[width=0.09\textwidth]{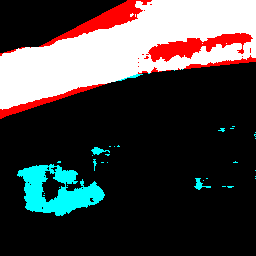} &\hspace{-4mm}
            \includegraphics[width=0.09\textwidth]{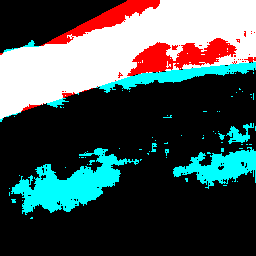} &\hspace{-4mm}
            \includegraphics[width=0.09\textwidth]{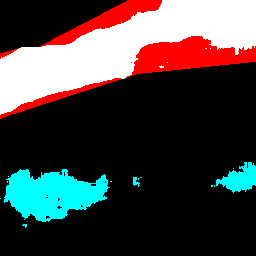} &\hspace{-4mm}
            \includegraphics[width=0.09\textwidth]{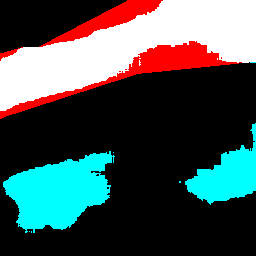} &\hspace{-4mm}
            \includegraphics[width=0.09\textwidth]{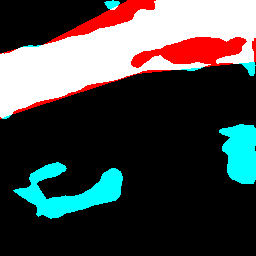} &\hspace{-4mm}
            \includegraphics[width=0.09\textwidth]{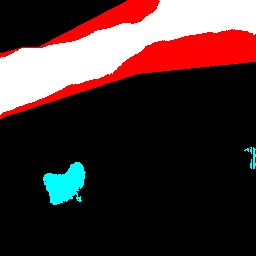} &\hspace{-4mm}
            \includegraphics[width=0.09\textwidth]{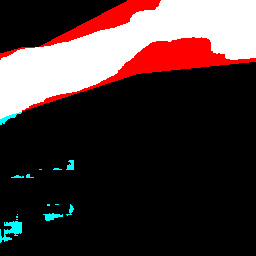} &\hspace{-4mm}
            \includegraphics[width=0.09\textwidth]{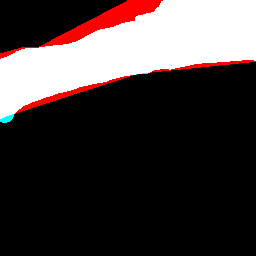} \\

            \includegraphics[width=0.09\textwidth]{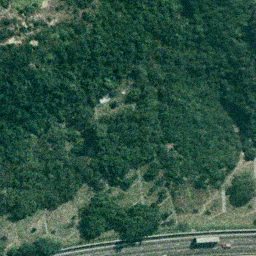} &\hspace{-4mm}
            \includegraphics[width=0.09\textwidth]{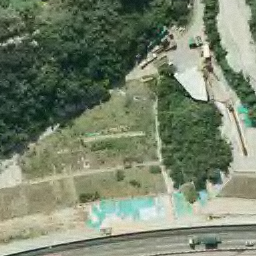} &\hspace{-4mm}
            \includegraphics[width=0.09\textwidth]{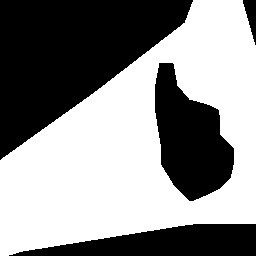} &\hspace{-4mm}
            \includegraphics[width=0.09\textwidth]{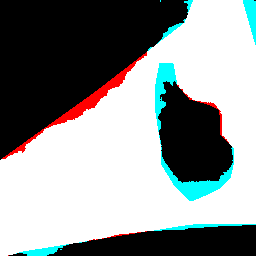} &\hspace{-4mm}
            \includegraphics[width=0.09\textwidth]{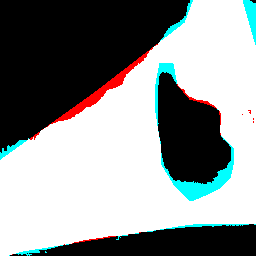} &\hspace{-4mm}
            \includegraphics[width=0.09\textwidth]{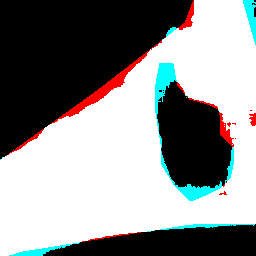} &\hspace{-4mm}
            \includegraphics[width=0.09\textwidth]{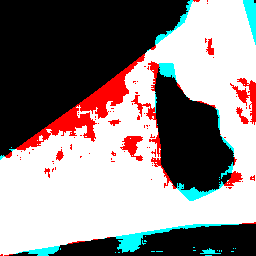} &\hspace{-4mm}
            \includegraphics[width=0.09\textwidth]{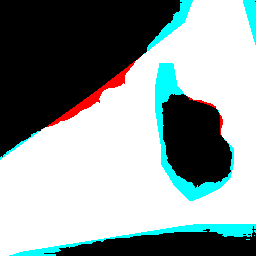} &\hspace{-4mm}
            \includegraphics[width=0.09\textwidth]{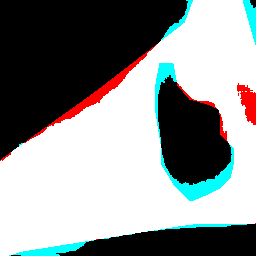} &\hspace{-4mm}
            \includegraphics[width=0.09\textwidth]{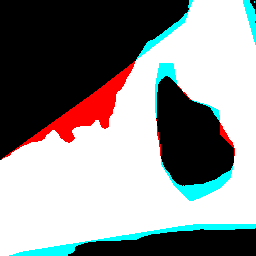} &\hspace{-4mm}
            \includegraphics[width=0.09\textwidth]{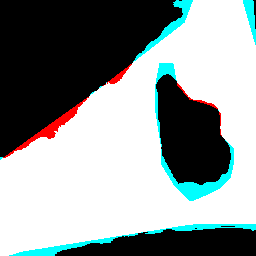} &\hspace{-4mm}
            \includegraphics[width=0.09\textwidth]{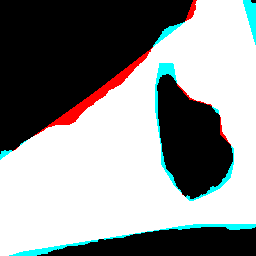} &\hspace{-4mm}
            \includegraphics[width=0.09\textwidth]{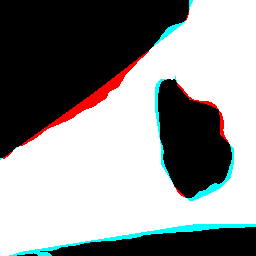} \\

            \includegraphics[width=0.09\textwidth]{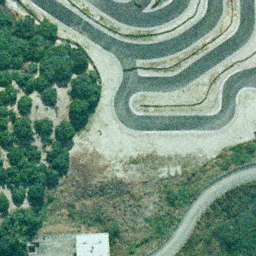} &\hspace{-4mm}
            \includegraphics[width=0.09\textwidth]{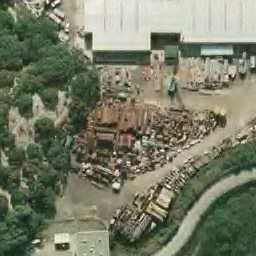} &\hspace{-4mm}
            \includegraphics[width=0.09\textwidth]{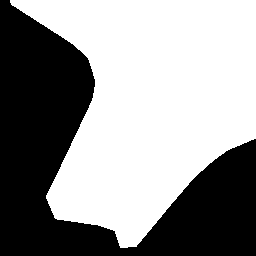} &\hspace{-4mm}
            \includegraphics[width=0.09\textwidth]{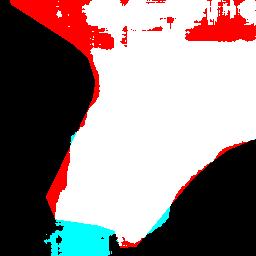} &\hspace{-4mm}
            \includegraphics[width=0.09\textwidth]{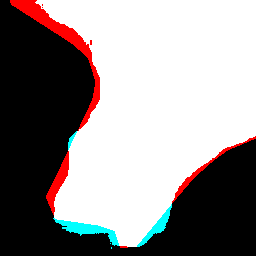} &\hspace{-4mm}
            \includegraphics[width=0.09\textwidth]{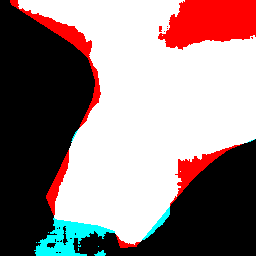} &\hspace{-4mm}
            \includegraphics[width=0.09\textwidth]{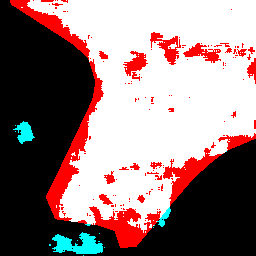} &\hspace{-4mm}
            \includegraphics[width=0.09\textwidth]{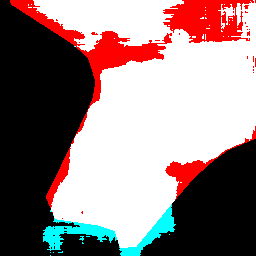} &\hspace{-4mm}
            \includegraphics[width=0.09\textwidth]{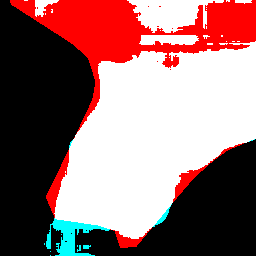} &\hspace{-4mm}
            \includegraphics[width=0.09\textwidth]{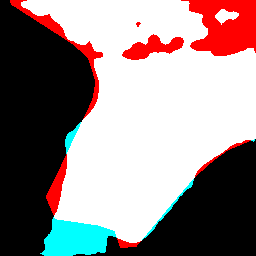} &\hspace{-4mm}
            \includegraphics[width=0.09\textwidth]{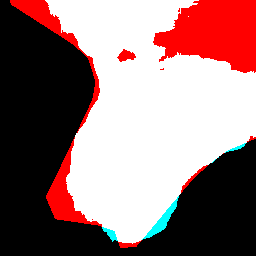} &\hspace{-4mm}
            \includegraphics[width=0.09\textwidth]{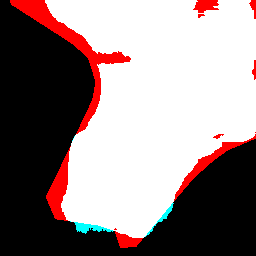} &\hspace{-4mm}
            \includegraphics[width=0.09\textwidth]{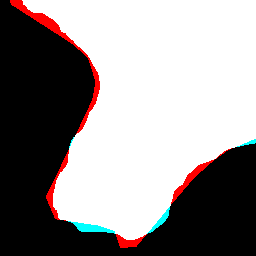} \\

            (a) & (b) & (c) & (d) & (e) & (f) & (g) & (h) & (i) & (j) & (k) & (l) & (m) \\
        \end{tabular}
    }
    \caption{Qualitative analysis of experimental results on the SYSU dataset~\cite{2021_IEEE_Geoscience_and_Remote_Sensing_Letters_DSAM}. (a) T1 images. (b) T2 images. (c) Ground truth. (d) FC-cat. (e) FC-ef. (f) FC-diff. (g) DSAM. (h) SNUNet. (i) L-Unet.  (j) BIT. (k) DSIFN. (l) P2V. (m) Ours. The displayed colors correspond to true positives (white), false positives (\colorbox{red!60}{red}), true negatives (black), and false negatives (\colorbox{my_cyan!40}{cyan}). }
    \label{fig:fig_SYSU}
\end{figure*}

\begin{figure*}[ht]
    \centering
    \resizebox{\textwidth}{!}{
        \begin{tabular}{@{}ccccccccccccc@{}}
            \includegraphics[width=0.09\textwidth]{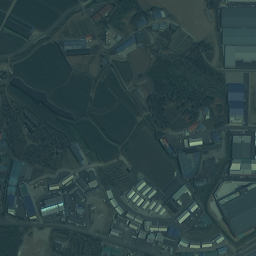} &\hspace{-4mm}
            \includegraphics[width=0.09\textwidth]{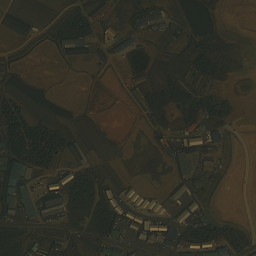} &\hspace{-4mm}
            \includegraphics[width=0.09\textwidth]{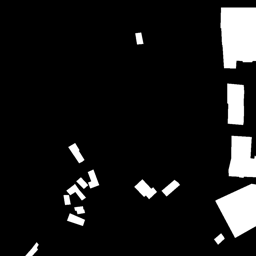} &\hspace{-4mm}
            \includegraphics[width=0.09\textwidth]{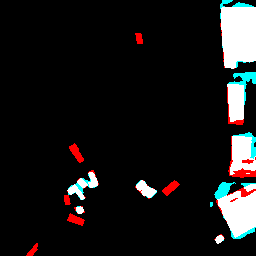} &\hspace{-4mm}
            \includegraphics[width=0.09\textwidth]{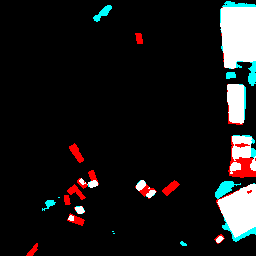} &\hspace{-4mm}
            \includegraphics[width=0.09\textwidth]{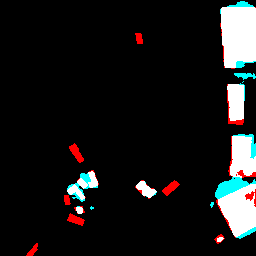} &\hspace{-4mm}
            \includegraphics[width=0.09\textwidth]{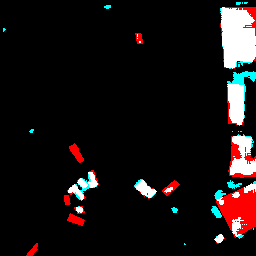} &\hspace{-4mm}
            \includegraphics[width=0.09\textwidth]{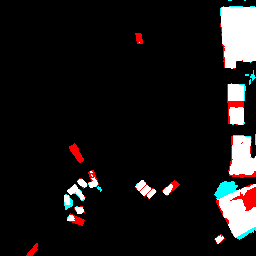} &\hspace{-4mm}
            \includegraphics[width=0.09\textwidth]{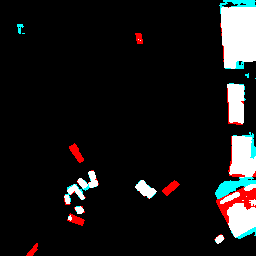} &\hspace{-4mm}
            \includegraphics[width=0.09\textwidth]{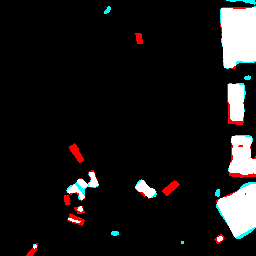} &\hspace{-4mm}
            \includegraphics[width=0.09\textwidth]{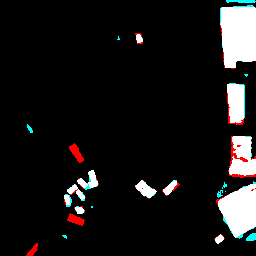} &\hspace{-4mm}
            \includegraphics[width=0.09\textwidth]{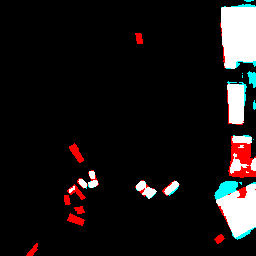} &\hspace{-4mm}
            \includegraphics[width=0.09\textwidth]{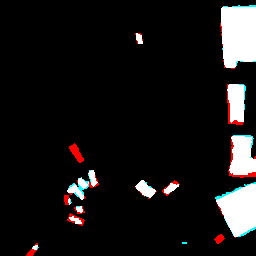} \\

            \includegraphics[width=0.09\textwidth]{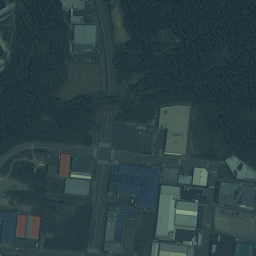} &\hspace{-4mm}
            \includegraphics[width=0.09\textwidth]{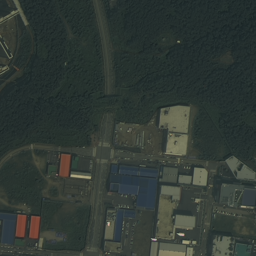} &\hspace{-4mm}
            \includegraphics[width=0.09\textwidth]{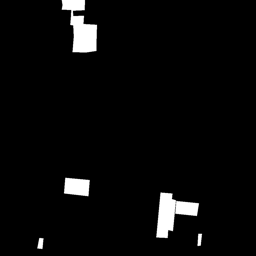} &\hspace{-4mm}
            \includegraphics[width=0.09\textwidth]{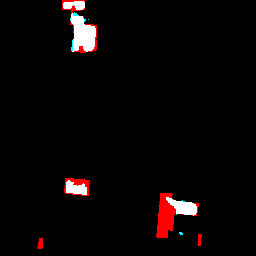} &\hspace{-4mm}
            \includegraphics[width=0.09\textwidth]{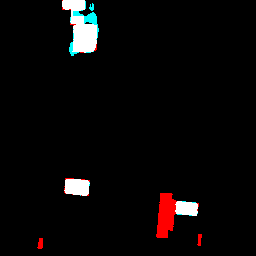} &\hspace{-4mm}
            \includegraphics[width=0.09\textwidth]{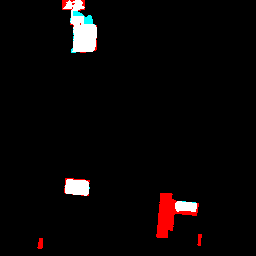} &\hspace{-4mm}
            \includegraphics[width=0.09\textwidth]{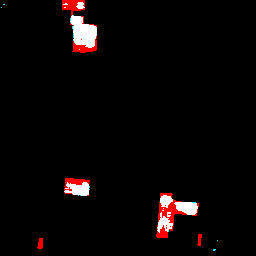} &\hspace{-4mm}
            \includegraphics[width=0.09\textwidth]{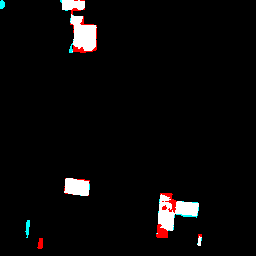} &\hspace{-4mm}
            \includegraphics[width=0.09\textwidth]{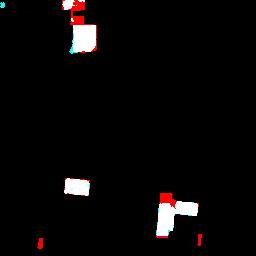} &\hspace{-4mm}
            \includegraphics[width=0.09\textwidth]{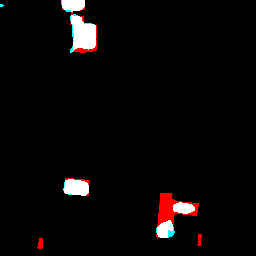} &\hspace{-4mm}
            \includegraphics[width=0.09\textwidth]{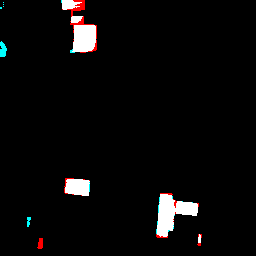} &\hspace{-4mm}
            \includegraphics[width=0.09\textwidth]{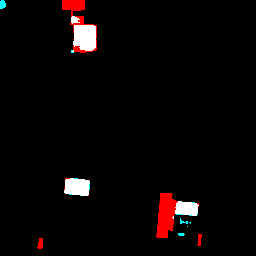} &\hspace{-4mm}
            \includegraphics[width=0.09\textwidth]{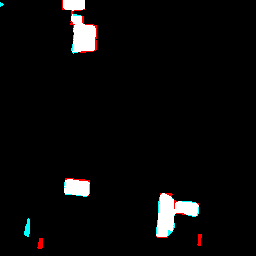} \\

            \includegraphics[width=0.09\textwidth]{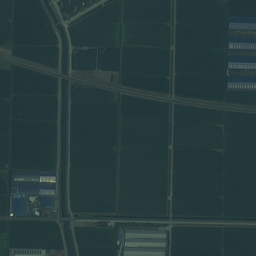} &\hspace{-4mm}
            \includegraphics[width=0.09\textwidth]{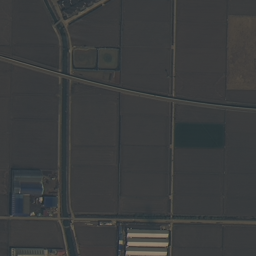} &\hspace{-4mm}
            \includegraphics[width=0.09\textwidth]{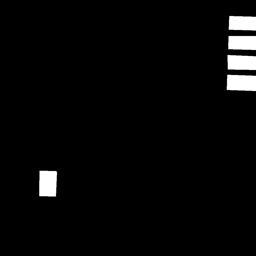} &\hspace{-4mm}
            \includegraphics[width=0.09\textwidth]{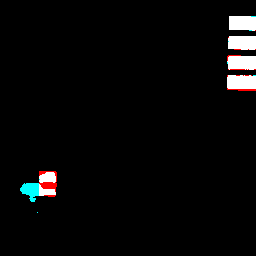} &\hspace{-4mm}
            \includegraphics[width=0.09\textwidth]{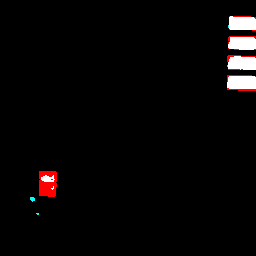} &\hspace{-4mm}
            \includegraphics[width=0.09\textwidth]{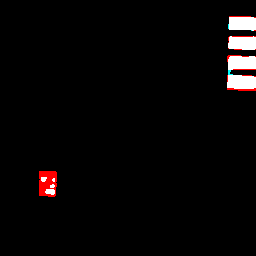} &\hspace{-4mm}
            \includegraphics[width=0.09\textwidth]{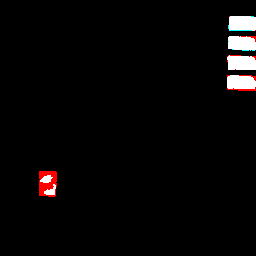} &\hspace{-4mm}
            \includegraphics[width=0.09\textwidth]{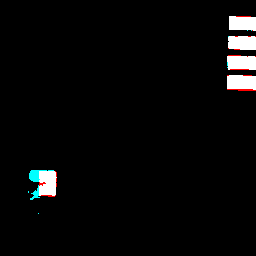} &\hspace{-4mm}
            \includegraphics[width=0.09\textwidth]{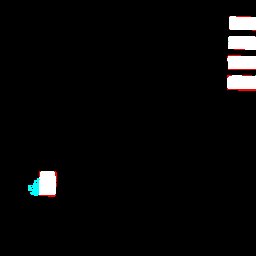} &\hspace{-4mm}
            \includegraphics[width=0.09\textwidth]{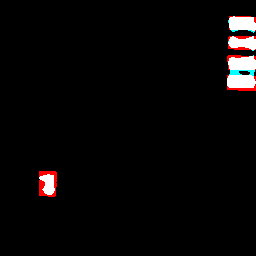} &\hspace{-4mm}
            \includegraphics[width=0.09\textwidth]{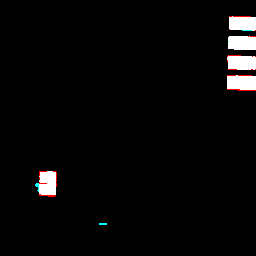} &\hspace{-4mm}
            \includegraphics[width=0.09\textwidth]{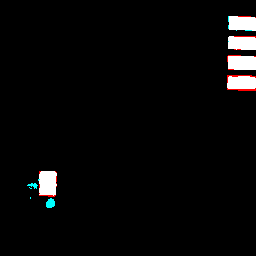} &\hspace{-4mm}
            \includegraphics[width=0.09\textwidth]{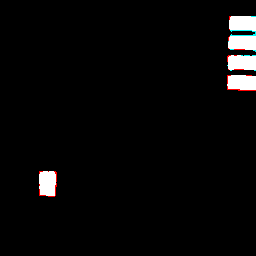} \\

            (a) & (b) & (c) & (d) & (e) & (f) & (g) & (h) & (i) & (j) & (k) & (l) & (m) \\
        \end{tabular}
    }
    \caption{Qualitative analysis of experimental results on the S2Looking dataset~\cite{37}. (a) T1 images. (b) T2 images. (c) Ground truth. (d) FC-cat. (e) FC-ef. (f) FC-diff. (g) DSAM. (h) SNUNet. (i) L-Unet. (j) BIT. (k) DSIFN. (l) P2V. (m) Ours. The displayed colors correspond to true positives (white), false positives (\colorbox{red!60}{red}), true negatives (black), and false negatives (\colorbox{my_cyan!40}{cyan}).}
    \label{fig:fig_S2Looking}
\end{figure*}

To comprehensively evaluate the effectiveness of our method, we select nine comparison methods and conduct visualizations to analyze each method's performance in depth. As shown in Figure~\ref{fig:fig_LEVIR}, on the LEVIR-CD+ dataset, our method detects change areas more completely compared to others, clearly identifying the boundaries and main structures of building changes. Figure~\ref{fig:fig_SYSU} demonstrates that our method performs better in detecting change areas, significantly reducing missed detections and false positives. Other methods exhibit unstable detection results under variations in lighting and noise, leading to numerous artifacts, while our method maintains high detection performance. As shown in Figure~\ref{fig:fig_S2Looking}, other methods perform poorly in edge detection, with vague outlines of change areas, whereas our method excels in capturing edge details and accurately identifying the boundaries of change areas. This superiority is primarily attributed to our method's ability not only to effectively learn the features of each channel but also to capture detailed changes, thereby enhancing the overall performance of the model.

Table~\ref{tab:tab2} presents the comparison results of different methods in terms of computational efficiency. In this table, the number of model parameters (Parms) is used to assess the spatial complexity of the model, while the number of floating-point operations per second (FLOPs) is used to measure the computational complexity of the model. Under unified parameter configurations, our method demonstrates outstanding computational efficiency. Specifically, in terms of parameters, our method reduces the number of parameters by approximately three times compared to the L-Unet method, approximately 0.85 times that of the BIT method, and significantly lower than the parameter counts of the DSIFN and SNUNet methods. In terms of computational load, the LCD-Net method maintains its lightweight characteristics, with the smallest computational load among the comparison methods, reducing the number of operations by approximately 18.2 times compared to the SNUNet method. Although the FC-diff method has lower parameters and computational loads, it fails to achieve the level of accuracy in change detection of the method proposed in this study. By considering both the accuracy and computational efficiency in change detection, our method exhibits a high-precision, lightweight model structure.

To maintain a lightweight structure, our method is designed with three modules that each retain lightweight characteristics. The TIF allows for channel exchanges while ensuring that gradients can smoothly backpropagate through other channels, without increasing the network's parameters or computational load. The FFM significantly reduces the number of parameters through simple convolution operations and basic computational processes. The GMM effectively achieves network lightweight with minimal parameters by calculating the \textit{L2}-norm and performing channel normalization. This design not only ensures computational efficiency but also enhances model performance.

\begin{table}[h]
\caption{Statistics on the computing efficiency of the model. (Color Convention:\textcolor{red}{best}, \textcolor{blue}{2rd best}) }
\centering
\renewcommand{\arraystretch}{2.1} 
\resizebox{\columnwidth}{!}{ 
\begin{tabular}{c@{\hskip 4pt}|@{\hskip 2pt}cccccccc}
\hline
Methods & FC-diff & DSIFN & DSAM & SNUNet & L-Unet & P2V & BIT & \textbf{LCD-Net} \\
\hline
FLOPs(G) & 10.62 & 164.56 & 65.68 & 246.22 & 34.63 & 40.50 & 16.88 & \textcolor{red}{4.45} \\
Parm(M) & \textcolor{red}{1.55} & 50.71 & 12.23 & 27.07 & 8.45 & 7.43 & 3.01 & \textcolor{blue}{2.56} \\
\hline
\end{tabular}
}
\label{tab:tab2}
\end{table}

\subsection{Ablation Experiments}
The innovations of our method mainly include three components: the TIF, the FFM, and the GMM. To validate the effectiveness of these three innovations, we conduct ablation experiments using the LEVIR-CD+, SYSU, and S2Looking datasets, selecting F1 and IoU as evaluation metrics. As shown in Table~\ref{tab:tab3}, all three modules can effectively enhance the performance of the main network with a relatively small increase in parameters. After adding the TIF, the F1-score improved by approximately 0.52\%, 1.95\%, and 1.45\% across the three datasets, while the IoU scores improved by approximately 0.53\%, 0.89\%, and 1.45\%, respectively. After adding the FFM, the F1-score improved by approximately 0.28\%, 0.53\%, and 0.78\% across the three datasets, and the IoU scores improved by approximately 0.47\%, 0.62\%, and 0.78\%, respectively. After adding the GMM, the F1-score improved by approximately 0.47\%, 0.36\%, and 0.51\%, and the IoU scores improved by approximately 0.80\%, 0.66\%, and 0.52\%, respectively. This is because each of the three modules achieves capabilities for feature activation, information fusion, and extraction of key features. Therefore, all three proposed modules can effectively enhance change detection performance.

\begin{table}[h!]
\caption{Ablation Study of the Main Modules of LCD-Net on Three Datasets}
\centering
\renewcommand{\arraystretch}{2.1} 
\setlength{\tabcolsep}{4pt} 
\resizebox{\columnwidth}{!}{ 
\LARGE 
\begin{tabular}{c c c c c|c c c c c c c c}
\hline
\multirow{2}{*}{} & Backbone & \multirow{2}{*}{TIF} & \multirow{2}{*}{FFM} & \multirow{2}{*}{GMM} & \multicolumn{2}{c}{LEVIR-CD+} & \multicolumn{2}{c}{SYSU} & \multicolumn{2}{c}{S2Looking} & \multirow{2}{*}{Parms(M)} & \multirow{2}{*}{Flop(G)} \\ 
& Network & & & & F1(\%) & IoU(\%) & F1(\%) & IoU(\%) & F1(\%) & IoU(\%) & & \\ 
\hline
(1) & \checkmark & & & & 90.21 & 82.50 & 78.36 & 66.21 & 56.51 & 39.39 & 2.04 & 3.98 \\ 
\hline
(2) & \checkmark & \checkmark & & & 90.73 & 83.03 & 80.31 & 67.10 & 58.00 & 40.84 & 2.04 & 3.98 \\ 
\hline
(3) & \checkmark & \checkmark & \checkmark & & 91.01 & 83.50 & 80.86 & 67.72 & 58.78 & 41.62 & 2.32 & 4.14 \\ 
\hline
(4) & \checkmark & \checkmark & \checkmark & \checkmark & 91.48 & 84.30 & 81.22 & 68.38 & 59.29 & 42.14 & 2.56 & 4.45 \\ 
\hline
\end{tabular}
}
\label{tab:tab3}
\end{table}

\begin{figure}
    \centering
    \includegraphics[width=\linewidth]{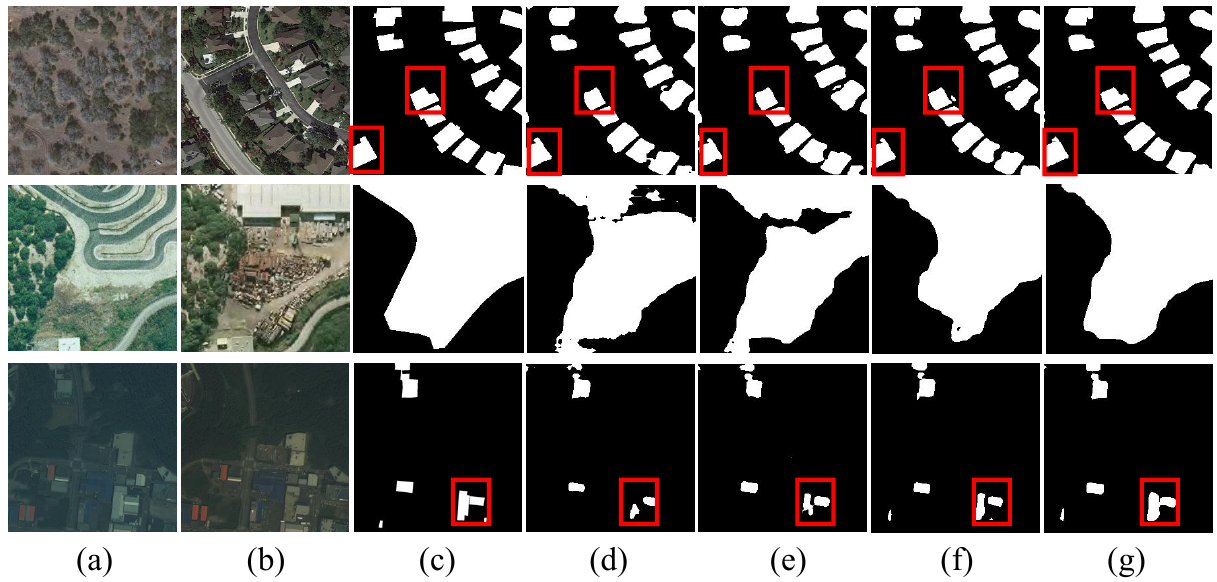}
    \caption{Ablation Study Results for the Proposed Method; (a): T1 image, (b): T2 image, (c): label map, (d): Backbone Network, (e): Backbone Network + TIF, (f): Backbone Network + TIF + FFM, (g): Backbone Network + TIF + FFM + GMM. }
    \label{fig:fig_Ablation_Study}
\end{figure}

To provide a more intuitive presentation of the effects of the three modules, Figure~\ref{fig:fig_Ablation_Study} displays the results of the ablation experiments, selecting a set of images from three datasets. Figure~\ref{fig:fig_Ablation_Study} (a)-(g) represent the images at time points T1 and T2, the label map, and the combinations of different modules (corresponding to (d)-(g) in Table~\ref{tab:tab3}). It is evident from the figure that when using the MobileNetV2 network as the backbone for prediction, some edges are not detected. This indicates that a single backbone network has limitations in capturing changed areas. By introducing TIF, the exchange of contextual information further activates the network features, effectively capturing more changed areas. After adding FFM, the model achieves significant improvement in the fusion of features from different layers, with some noise being suppressed, thus enhancing prediction accuracy. With the incorporation of GMM, the model not only excels at extracting key information but also captures edge details more precisely, significantly enhancing overall performance.

\section{Conclusion}
This paper proposes a lightweight and efficient remote sensing change detection model, LCD-Net, which aims to reduce model parameters while enhancing change detection accuracy. 
By introducing the Temporal Interaction and Fusion module (TIF), the model enhances the interaction between dual-temporal features, improves contextual awareness, and maintains the lightweight characteristics of the model. 
The Feature Fusion Module (FFM) deepens the integration of multi-scale features, increasing sensitivity to subtle changes and improving the model's change detection capability in complex backgrounds. 
The Gated Mechanism Module (GMM) in the decoder dynamically adjusts channel weights, further strengthening the extraction of key features and reducing the impact of redundant information on detection accuracy. 
Experimental results demonstrate that LCD-Net achieves good performance across various remote sensing change detection tasks while maintaining low computational overhead. 
The model not only excels in handling complex backgrounds and subtle changes but also provides an effective solution for practical applications such as remote sensing monitoring and environmental change assessment.

\bibliography{main} 
\end{document}